\documentclass[10pt,twocolumn,letterpaper]{article}

\usepackage{iccv}
\usepackage{times}
\usepackage{epsfig}
\usepackage{graphicx}
\usepackage{amsmath}
\usepackage{amssymb}
\usepackage{xspace}
\usepackage{multirow}
\usepackage{enumitem}
\usepackage{hhline}
\usepackage{xcolor}
\usepackage{ulem}
\usepackage{enumitem}
\usepackage{comment}
\usepackage{makecell}
\usepackage{listings}
\usepackage{algorithm} 
\usepackage{algpseudocode} 
\usepackage{algorithmicx}
\usepackage{amsmath}
\usepackage{pdfpages}
\usepackage{mwe}

\usepackage[breaklinks=true,bookmarks=false]{hyperref}

\newcommand{\sysname}{ODIN\xspace}

\newcommand{\jihoon}[1]{\textcolor{red}{\textbf{Jihoon:} #1}}

\iccvfinalcopy 

\def\iccvPaperID{****} 

\ificcvfinal\fi

\begin{document}

\title{ODIN: \underline{O}n-demand \underline{D}ata Formulation to M\underline{I}tigate Dataset Lock-i\underline{N}}

\author{
 SP Choi, Jihun Lee, Hyeongseok Ahn, Sanghee Jung, Bumsoo Kang\\
 Lotte AI Research\\
 {\tt\small \{seungpyo.choi, jihoon8798, hyeongseok\_ahn, sanghee.jung, bumsoo.kang\}@lotte.net}
 }

\maketitle
\ificcvfinal\thispagestyle{empty}\fi

\begin{abstract}
   \sysname is an innovative approach that addresses the problem of dataset constraints by integrating generative AI models. Traditional zero-shot learning methods are constrained by the training dataset. To fundamentally overcome this limitation, \sysname attempts to mitigate the dataset constraints by generating on-demand datasets based on user requirements.
   \sysname consists of three main modules: a prompt generator, a text-to-image generator, and an image post-processor. To generate high-quality prompts and images, we adopted a large language model (e.g., ChatGPT), and a text-to-image diffusion model (e.g., Stable Diffusion), respectively. We evaluated \sysname on various datasets in terms of model accuracy and data diversity to demonstrate its potential, and conducted post-experiments for further investigation.
 Overall, \sysname is a feasible approach that enables AI to learn unseen knowledge beyond the training dataset.
\end{abstract}
\vspace{-1em}

\section{Introduction}

The primary drawback of traditional supervised machine learning is its dependency on the training dataset for acquiring knowledge. This means that a model designed for a particular classification task is unable to classify classes that were not included in the dataset. 
Researchers have been exploring ways to enable machine learning techniques to obtain \textit{learning-ability for new things}, similar to how humans learn.
Zero-shot learning~\cite{larochelle2008zero} is a popular approach in this direction, which tackles the challenge of handling unseen classes not present in the training dataset. 
Embedding-based zero-shot learning~\cite{annadani2018preserving, lampert2009learning, palatucci2009zero} enables a model to represent unseen classes as a combination of features solely learned from the training data, leading to a strong dependency on the dataset.
On the other hand, generative zero-shot learning~\cite{gao2020zero, lin2021zstgan, narayan2020latent, ramesh2021zero, xian2018feature, xian2019f,  zhao2022boosting} is the method that trains a generative model to generate latent features for unseen classes based on the training dataset. The latent features are integrated with the training dataset at the feature level, allowing the model to learn the knowledge of the unseen classes. However, since the generative model is trained on the training dataset, the dependency on the dataset still persists. 
In conclusion, both zero-shot learning methods are still fundamentally constrained by the prior knowledge of the training dataset and may not perform well on unseen classes in completely different domains.
 
In this study, we address the challenge of dataset lock-in, which is being constrained by the dataset, in a different way. While an ideal solution would be creating a large dataset that contains all the necessary knowledge, constructing such a dataset and training a model on it may not be practical due to its massive learning capacity requirement. 
Previous attempts at tackling such challenges focused on utilizing prior knowledge to deal with new classes, which does not fundamentally solve the underlying problem of being locked in the dataset. 
As an alternative, we attempt to mitigate the dataset constraints by generating on-demand datasets based on user requirements. In this light, we propose \sysname, a system that dynamically formulates customized datasets for users by leveraging text-to-image generation techniques.

\begin{figure*}[ht]
\centering
\includegraphics[width=0.89\linewidth]{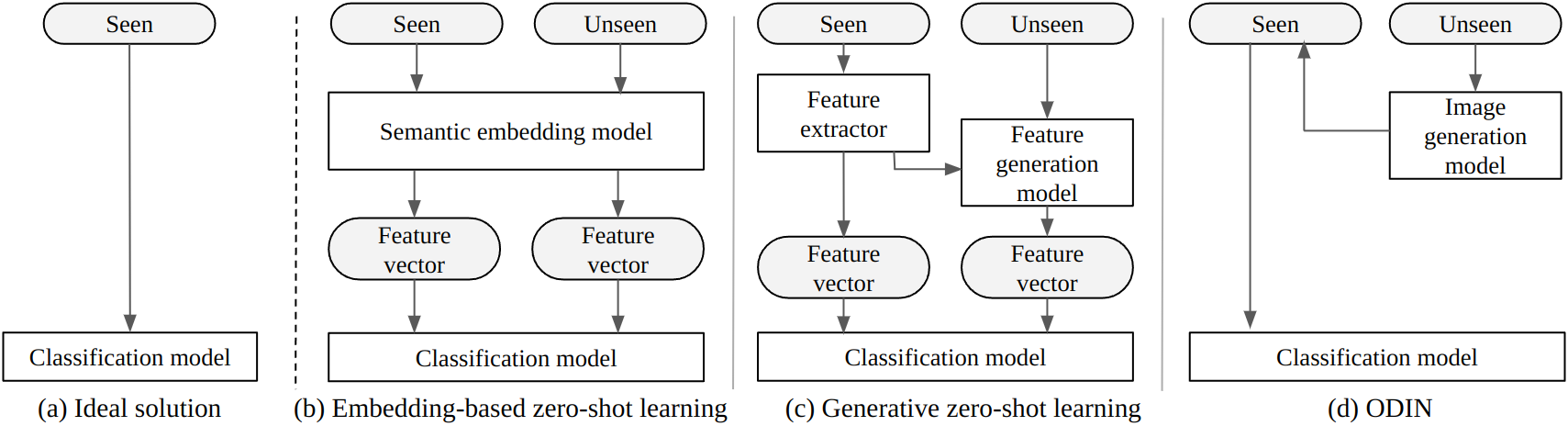}
\caption{\sysname vs. zero-shot learning, where the given task is to classify unseen labels.
}
\vspace{-1.3em}
\label{fig:nuo_intro}
\end{figure*}

Figure~\ref{fig:nuo_intro} highlights the key structural difference between \sysname and typical zero-shot learning approaches, where the main task is to classify unseen classes.

\begin{itemize}[noitemsep, topsep=0pt, leftmargin=*]
  \item The ideal solution in (a) is that all the data and labels required to train a classification model are already present in the dataset. In other words, \textit{unseen} data does not actually exist, which eliminates the need for zero-shot learning techniques (ideal but not realistic).
  
  \item Embedding-based zero-shot learning~\cite{annadani2018preserving, lampert2009learning, palatucci2009zero} in (b) utilizes a semantic embedding model that learns the seen data and converts semantic attributes into feature vectors. This approach allows the visual feature of the image and the semantic feature of the class to be projected to the same latent space. Then, it recognizes the unseen class based on the similarity between the feature vectors of seen and unseen data.
  
  \item Generative zero-shot learning~\cite{gao2020zero, lin2021zstgan, narayan2020latent, ramesh2021zero, xian2018feature, xian2019f,  zhao2022boosting} in (c) generates latent feature vectors for unseen labels. Firstly, the feature extractor converts the pairs of image-label (i.e., training dataset) into the pairs of feature-label (i.e., feature vectors). The feature generation model is then trained using these feature vectors to generate additional feature vectors for the unseen labels. These converted and generated feature vectors are integrated to train the classification model.
  
  \item As in (d) for \sysname, it dynamically formulates an on-demand dataset by directly generating image data from the unseen labels without relying on the training dataset and incorporating them with the existing dataset. From the model's perspective, the previous unseen classes become seen classes as the unseen data are now included in the dataset.
  This approach differs from generative zero-shot learning (c) in the way of handling unseen classes. First, the data generation model in \sysname (i.e., image generation model) is independent of the training dataset (i.e., seen classes), while the one in the generative zero-shot learning relies on the dataset. Second, \sysname handles the unseen classes at the data level (i.e., image-label), while zero-shot learning handles them mostly in the feature level (i.e., feature-label). These fundamental differences mitigate the dataset constraints and enable to create new classes, even if they were not in the original dataset.
\end{itemize}

To evaluate the feasibility of \sysname, we compare the performance of the model trained on the \sysname dataset with the one trained on the conventional dataset. 
We followed a common training hyperparameter setting in the experiment, which is likely already tuned to the conventional datasets.
Since \sysname-generated images and real images may differ in various perspectives, there could be other optimization settings for \sysname. However, hyperparameter optimization was out of our scope in this study.

\section{Related Work}

\sysname can be viewed as an advanced approach that addresses multiple research problems, including data augmentation/generation and zero-shot learning for unseen classes. We discuss the previous angles and approaches taken toward each problem, and position \sysname among them.

\subsection{Data augmentation and generation}

Research on data augmentation has become a critical aspect in deep learning field, since deep neural networks heavily rely on the size of training data. 
Data augmentation aims to artificially inflate the size of a dataset while keeping the labels fixed, spanning from common methods like rotating, flipping, and cropping, to more advanced methods like mixup~\cite{zhang2017mixup}, cutmix~\cite{yun2019cutmix},  UniformAugment~\cite{lingchen2020uniformaugment} and TrivialAugment~\cite{muller2021trivialaugment}. Nevertheless, such augmentation techniques require a minimum amount of data to be performed, which is dependent on task and domain.

On the other hand, recent advances in image generation have been gaining attention as a breakthrough for the long-standing problem of insufficient training data. Ravuri and Vinyals~\cite{ravuri2019seeing} implement BigGAN~\cite{brock2018large} as a large scale, high-fidelity image generative model. Their conclusion is that the mere addition of generated samples results in a lengthy training time with a minor improvement. Several attempts have been introduced to overcome such performance issues in investigating learning strategies~\cite{besnier2020dataset} or generative models~\cite{he2022synthetic, zhang2021datasetgan}. These efforts aimed at training a model with generated images, assuming that a gap between real and generated images is the reason for the poor performance.


Many attempts implemented state-of-the-art image generation models to reduce this gap by improving the quality of generated images. In contrast, \sysname takes a different approach by reducing the gap in the feature space, which is inserting pixel noise to the generated images. While this may seem to manipulate the images more \textit{unrealistic}, note that the key idea is to reduce the distance between the real and generated images in the feature space.

\subsection{Zero-shot learning for unseen label}
Zero-shot learning~\cite{larochelle2008zero} refers to the capability of a model to classify objects that were not presented in the training dataset by transferring knowledge from seen classes to unseen classes.
A number of proposed approaches enabling models to recognize unseen classes without prior knowledge of and access to unseen data. These can be broadly categorized into two approaches, embedding-based zero-shot learning and generative zero-shot learning.

Embedding-based zero-shot learning~\cite{annadani2018preserving, lampert2009learning, palatucci2009zero} firstly trains a model to embed semantic attributes of images, and maps them to classes in a latent space to find class representation.
By this means, it is possible to recognize unseen classes based on the similarity between the representation of the image to be recognized and those of the class.
The advantage of this approach is that it is easy to implement and relatively requires low computing resources. 
However, in generalized zero-shot learning~\cite{chao2016empirical} (i.e., recognizing not only unseen but also seen), 
the model has a risk to be biased when the images of unseen classes are insufficient, and eventually get tendency to recognize only the seen classes.

Generative zero-shot learning emerged with the development of deep learning-based generation models. The idea is to convert zero-shot learning into supervised learning by generating images~\cite{ramesh2021zero} or features~\cite{xian2018feature, zhao2022boosting}.
However, due to the unpleasant performance of existing image generation models, most of the studies are directed toward to generating features rather than images themselves. This feature generation approach overcomes the bias problem that occured in embedding-based methods, as it can generate a large number of samples for unseen classes~\cite{pourpanah2022review}. However, there is a potential generalization problem on unseen classes because the generative model is only trained on seen data. 


In other words, zero-shot learning methods that attempt to overcome the drawback of supervised learning (i.e., dataset lock-in) also fundamentally suffer from the same drawback.
In contrast, \sysname directly formulates datasets that are independent of existing datasets and aligned with the user's requirements, which fundamentally solves the problem of dataset lock-in. Furthermore, \sysname has the potential to become the foundational approach for ideal supervised learning that is free from the data constraints.
\section{\sysname Design} 
\begin{figure}
\centering
\includegraphics[width=0.8\linewidth]{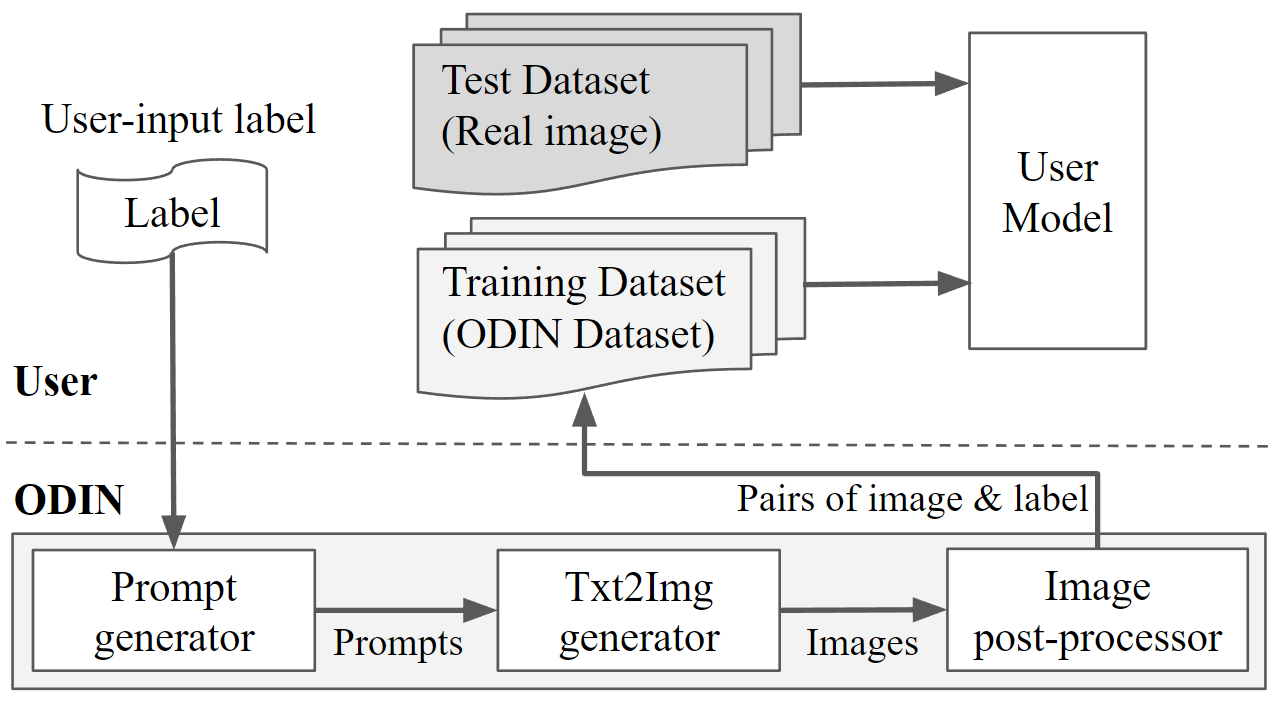}
\caption{The process of dynamic dataset formulation in \sysname.}
\vspace{-1em}
\label{fig:nuo_overview}
\end{figure}

In this section, we present an overview of the design process behind \sysname, which aims to dynamically formulate on-demand dataset based on the user request, namely dynamic data formulation. \sysname consists of three main components, including prompt generator, text-to-image generator, and image post-processor.
Figure~\ref{fig:nuo_overview} illustrates the overall process of dynamic data formation in \sysname. First, the prompt generator receives a label from the user and generates a prompt.
Second, a set of images is generated by the text-to-image (denoted txt2img, hereafter) generator based on the given prompt.
Finally, the image post-processor reduces the invisible gap that may exist between the generated and the real images, although the generated images seem realistic to the human eye.
The output images generated by \sysname are then used as input to train the models.
We name this newly generated dataset as \textbf{\sysname dataset}. 

\begin{figure*}
\centering
\includegraphics[width=0.93\linewidth]{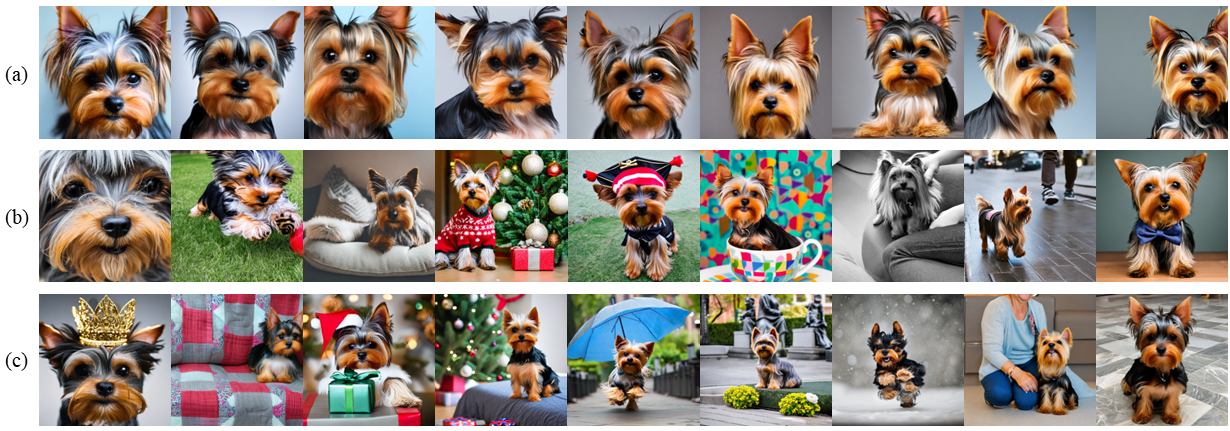}
\caption{Images generated from \sysname. (a) images generated from the naive prompt in the initial version (Iteration 1), (b) images generated from the ChatGPT-based prompt generator (Iteration 2), (c) images generated from blip-v2-based prompt extraction (Iteration 2).}
\vspace{-0.5em}
\label{fig:incereased_variance_with_chatGPT}
\end{figure*}


To begin the design process, we carefully selected a suitable dataset that would meet our criteria for fine-grained classification with a sufficient number of classes and a relatively high image resolution.
Our choice was Oxford-IIIT Pet dataset (denoted Oxford Pet, hereafter)~\cite{parkhi12a}, which includes 37 classes of pet images, to list a few, Bengal, Egyptian Mau, Pomeranian, etc.
The dataset contains images of varying resolutions, with an average of 430x390, and around 200 images included in each class.
Then, we iteratively designed each component of \sysname based on the classification task on the Oxford Pet.
To make decisions in each step, we evaluated the performance of Swin-v2 (started from Imagenet-1k checkpoint), using a common training parameter setting to focus solely on the dataset itself rather than parameter optimization.
The hyperparameter setting is: learning rate=2e-5, optimizer=adam, loss=cross-entropy, batch size=50, without any additional lr-scheduling. We report the best accuracy within 50 epochs.





\subsection{Iteration 1: Initial design}
As an initial step, we designed basic components of \sysname with minimal guidance in order to assess its feasibility. Since the primary goal of \sysname is to generate images, we incorporated Stable Diffusion-v2~\cite{rombach2021highresolution}, a publicly available txt2img generation model. Subsequently, we initially designed \sysname with a prompt generator and an image post-processor on top of the txt2img generator.
\begin{itemize}[noitemsep, leftmargin=*]
    \item \textbf{Prompt generator}: It generates simple prompts in a style of ``\textit{a photo of one \{class\} \{context\}}", where \{class\} represents the user-provided label and \{context\} is the word `\textit{pet}', as the design process is based on the Oxford Pet. The reason we added the `\textit{pet}' to the prompt is to avoid homophone errors in some classes (e.g., boxer).
    \item \textbf{Txt2Img generator}: It uses Stable Diffusion-v2, a generalized pre-trained txt2img generation model released under an open-source license by Stability AI, taking the output of the prompt generator as input.
    \item \textbf{Image post-processor}: It applies Gaussian blur with a kernel size of 5, which is a widely used technique for image noise reduction.
\end{itemize}


Using the images generated by the initial version of \sysname, we achieved a classification accuracy of \textbf{78.6\%} on the Oxford Pet, which seems a decent result for a first attempt, yet still identified rooms for improvement.
To improve the accuracy further, we examined the generated images. We observed that they appeared less diverse compared to those in the real dataset. For example, Figure~\ref{fig:incereased_variance_with_chatGPT}(a) shows a set of images of `\textit{yorkshire-terrier}' class generated by the initial version of \sysname. These images have similar colors, backgrounds, and compositions. We concluded that increasing the diversity in the generated images is our next priority.

\subsection{Iteration 2: Prompt generator}
In the recent advances in image generation, we design \sysname to be flexible enough to accommodate a number of image generation models. To this end, we tried to keep the image generation module untouched so that it could be replaced with other models in the future. Instead, we focused on addressing the diversity issue at the prompt level. We developed the prompt generator to generate various prompts as inputs for the txt2img generator, aiming to finally generate diverse images while maintaining the flexibility of the image generation module. 
To develop the prompt generator, we tested two different approaches: 1) generating prompts with pre-trained large language models (LLMs) and 2) extracting prompts from real images with pre-trained image caption LLMs. We performed the classification task with changing only the prompt generator to observe the suggested strategies respectively.

\smallskip
\noindent \textbf{Prompt generation}: 
For the generative language model, we chose ChatGPT~\cite{chatgpt} which is a text-based interaction service developed on GPT-3.5 by OpenAI. To generate the prompts for each class, we sent queries to ChatGPT with the following input: ``\textit{Can you recommend 10 simple prompts for image creation? I want to generate photo-realistic {class} images with txt2img model}".
We observed that the ChatGPT-suggested prompts generated much more diverse images than the naive prompts.

\smallskip
\noindent \textbf{Prompt extraction}:
To obtain diverse prompts by extracting them from real images, we adopted blip-v2 model (blip2-opt, pretrained-opt2.7b)~\cite{li2023blip}, a recently developed high-performance image captioning model (hits 96.9\% accuracy in img2txt captioning under the Flickr30K dataset)
However, even with a good image captioning model, biased or low-quality source images would lead to insufficient quality of the captioned prompts for generating a good training dataset.
To avoid such an issue, we use Oxford Pet training dataset for captioning. Overall, blip-2 image captioning works well; however, the captioned description of the subject tends to be coarse compared to the labels in the dataset. For example, an image of `\textit{yorkshire-terrier}' is captioned as an image of `\textit{dog}'. Therefore, in reverse, we replace the word of coarse subjects with the labels of the dataset (e.g., replacing `\textit{dog}' with `\textit{yorkshire-terrier}'). Since we directly used Oxford Pet, this may represent the peak performance of image-captioning under optimal settings.



\begin{table}
\centering
\small
\begin{tabular}{|c||c|c| }
\hline
 \textbf{Prompt generator} & \textbf{Top-1} (\%) & \textbf{Top-3} (\%) \\
\hline
 Naive prompt & 78.6 & 93.9 \\
 ChatGPT-based generation & 85.4 & 95.5 \\   
 blip-v2-based extraction & 85.8 & 96.8 \\    
  \hline
 \end{tabular}
 \caption{The accuracy of applying prompt generator.}
 \vspace{-1em}
\label{section3-prompt}
\end{table}

Figure~\ref{fig:incereased_variance_with_chatGPT} shows sampled generated images using (a) the naive prompt in our initial design, (b) prompts suggested by ChatGPT, and (c) prompts extracted from the training dataset using blip-v2. We observed that ChatGPT-suggested prompts and blip-v2 based extracted prompts generate more diverse images than the naive prompts. As for the image captioning in (c), we provide examples of the source images in the training dataset and the generated images of captioned prompts from the source images in our supplementary.

Table~\ref{section3-prompt} presents the results of each method. ChatGPT-based prompt generation achieved 85.4\%, while blip-v2 based prompt extraction achieved 85.8\% accuracy. Both approaches demonstrate a significant improvement compared to the naive prompt generator.
This indicates that the quality of prompts is heavily related to the quality of dataset, followed by the model performance.
Although there is no significant difference between the two approaches, they have different practical implications.
The prompt extraction requires source images to be captioned while the prompt generator does not. 
In our design study, we used Oxford Pet as the source images, since it was the dataset we tested.
However, using training set of a specific dataset as the source images and captioning them as prompts are difficult to generalize.
In contrast, ChatGPT-based prompt generation is easy to incorporate into the \sysname pipeline, as the output from ChatGPT can be directly used as a prompt.
With these considerations, we decided to use the ChatGPT-based prompt generator for \sysname.



\begin{table}
\centering
\small
\begin{tabular}{|c||c|c| }
\hline
 \multirow{2}{*}{\textbf{Method}} & \textbf{Accuracy} (\%) & \textbf{Accuracy} (\%) \\
 & training dataset only & training\&test dataset \\
\hline
 Gaussian blur & 85.4 & \textbf{85.3}\\
 Gaussian noise & 85.1 & 81.5\\   
 Localvar noise & 85.3 & 82.8\\    
 Poisson noise & 83.2 & 82.3\\      
 Salt noise & \textbf{86.4} & 81.5\\
 Pepper noise &  83.6 & 82.0\\
 S\&P noise & 84.7 & 82.8\\
 Speckle & 84.6 & 83.0 \\
 \hline
 \end{tabular}
 \caption{The accuracy results on applying image post-processing. The post-processing were applied only for the training dataset (left) and the test dataset as well (right).}
 \vspace{-1em}
\label{section3-table}
\end{table}

\begin{table*}
\centering
\small
\begin{tabular}{|c||c|c||c|c||c|c||c|c|}
\hline
\multicolumn{9}{|c|}{Top-1 accuracy test w/ and w/o applying image post-processing (\%)} \\ 
\hline
Test dataset & \multicolumn{2}{c||}{Oxford Pet} & \multicolumn{2}{c||}{Caltech} &  \multicolumn{2}{c||}{Indian Food}  & \multicolumn{2}{c|}{CIFAR-100}\\ 
\hhline{|=||==||==||==||==|}
\textbf{Trained network} & \textbf{None} & \textbf{Best} & \textbf{None} & \textbf{Best} & \textbf{None} & \textbf{Best} & \textbf{None} & \textbf{Best} \\
\hline
ResNet-152 & 81.5 & 85.8 (Ln) & 86.5 & 86.5 (None) & 66.4 & 70.2 (Ln) & 60.1 & 60.8 (Gn) \\
\hline
ResNeXt101 & 85.1 & 87.0 (Ln) & 89.6 & 91.4 (Sn) & 72.3 & 72.3 (None) & 61.2 & 61.6 (Ln) \\ 
\hline
ViT-b-16 & 82.4 & 83.7 (Sn) & 86.0 & 86.0 (None) & 71.2 & 73.7 (Pn) & 61.1 & 61.1 (None) \\ 
\hline
Swin-v2 & 85.4 & 86.4 (Sn) & 86.4 & 88.0 (Sn) & 71.8 & 72.7 (Sp) & 61.4 & 61.5 (Sp) \\
\hline
RegNet & 84.4 & 84.5 (Gn) & 85.1 & 88.4 (Ln) & 71.3 & 71.5 (Sp) & 53.4 & 55.3 (Gn)\\ 
\hline
\end{tabular}
 \caption{The results on accuracy tests with and without applying image post-processing. Post-processing techniques are denoted as follows; Ln: Localvar noise, Sn: Salt noise, Gn: Gaussian noise, Pn: Poisson noise, and Sp: Speckle.}
 \vspace{-1em}
 \label{section4-table:1} 
\end{table*}

\subsection{Iteration 3: Image post-processor}
Despite generating more diverse images, the accuracy with \sysname was still lower than that of real image dataset.
Previous research discussed in \S2.1 investigated such problems and concluded that there is a gap between the real images and generated ones. 
They addressed it by increasing the image quality closer to real images. However, we approached the problem differently: reducing the distance in the feature space.
We hypothesized that adding noises to the generated images could smooth out the specific features of the generated images.
We experimented with several noise functions on the model with ChatGPT-based prompts (from the \S3.1) as the base (i.e., the base accuracy: 85.4\%). Table~\ref{section3-table} (left) presents the accuracy of the model with each noise function. While the salt noise performed the best with 86.4\% (the increase of 0.7\% from the base), we also observed that most of the functions show similar results (M=84.79, SD=0.95). Since the experiment was conducted on Oxford Pet under a specific training condition, there is a possibility that the performance of the noise functions may significantly vary depending on the task or model architecture.



We also hypothesized that applying post-processing to real images in the test dataset could smooth out their specific features. We expected such an attempt would reduce the gap between the generated and the real images in the feature space, leading to a better classification performance. To test our assumption, we applied the same post-processing to the test dataset during the inference. Contrary to our expectations, the result was slightly less than that of applying to the training dataset only (M=82.65, SD=1.14), as shown in Table~\ref{section3-table} (right).
Our interpretation is that the generated images have specific features that differ from those of real images, and applying post-processing smooths out such differences. However, applying post-processing to real images acts as noise and reduces the performance of the model. 
Therefore, we conclude to apply post-processing only to the generated images in the training dataset.

\subsection{Dynamic dataset formulation}
We built the first prototype of \sysname by implementing the prompt generator and the image post-processor as we designed. To evaluate the feasibility of dynamically formulating on-demand datasets, we conducted small-scale Proof of Concept (PoC) experiment. Thorough experiments on other datasets are discussed in the next section.

For PoC experiment, we conducted 101 rounds of tests. In each round, we created a test dataset randomly selected and combined 10 classes from each of the Oxford Pet and Indian Food, resulting in a total of 20 classes.
We ensured the number of data for each class was balanced by randomly sampling images with the minimum number of data in a class.
As for the training dataset, based on the selected labels, we generated 10 different prompts for each class using ChatGPT, followed by generating 18 images from each prompt. In total, 180 images for each class were generated as the training set.
Overall, we achieved an average accuracy of \textbf{88.4\%} with a SD of 3.25 across the 101 rounds. 
The result demonstrated the potential of \sysname for dynamic dataset formulation.
Note that \sysname has, technically, the capability to generate an unlimited amount of data. 



\section{Technical Probe}

\subsection{\sysname Evaluation}
We conducted evaluations on a number of tasks to examine the quality of \sysname dataset. We focused on two main aspects: \textbf{accuracy} and \textbf{diversity}. We measured the accuracy of various models trained on \sysname dataset under different datasets. As for diversity, we compared the variance of structural similarity index (SSIM) between real datasets and those generated by \sysname.

\vspace{-0.8em}
\subsubsection{Accuracy}
\vspace{-0.2em}

In our design process in \S3, we conducted several experiments using a fixed dataset (i.e., Oxford Pet) and model structure (i.e., Swin-v2) to verify the feasibility of \sysname. To evaluate the final design of \sysname, we measured the accuracy with various datasets and models. 
We used four datasets in our evaluation, including Oxford Pet~\cite{parkhi12a}, Caltech~\cite{FeiFei2004}, Indian Food~\cite{rajiIndian}, and CIFAR-100~\cite{CIFAR100}.
We used all classes of the datasets except for the `\textit{Face-easy}' class (a cropped version of `\textit{Face}' class) of Caltech. We omitted it due to its loose connection between the label and the photo, resulting in the difficulty of generating proper image with \sysname. 
As for the model structures, we used five models, including ResNet-152~\cite{he2016deep}, ResNeXt101~\cite{xie2017aggregated}, ViT-b-16~\cite{dosovitskiy2020image}, Swin-v2~\cite{liu2022swin}, and RegNet~\cite{radosavovic2020designing}. 
We applied the hyperparameters globally as described in \S3. 
In the experiments, we generated \sysname dataset in 768x768 resolution, which appeared to be the optimal resolution for Stable Diffusion-v2. We resized the images to 224x224 resolution before feeding them to the models, and began the training from the Imagenet-1k checkpoint. Table \ref{section4-table:1} shows the results.


Focusing on the accuracy of the models for each dataset, we found that \sysname performs well across various model structures without tuning the parameters, not just on the model we used for our design (i.e., Swin-v2).
The top-1 best accuracy also did not show significant variance across models for each dataset, including Oxford Pet (M=83.76, SD=1.54, Caltech (M=86.72, SD=1.52), Indian Food (M=70.6, SD=2.14), and CIFAR-100 (M=63.88, SD=2.24). 
The results demonstrate the feasibility of \sysname for various model structures. 
However, we observed that the accuracy of the model for CIFAR-100 tends to be lower than that of the others, possibly due to the low resolution of CIFAR-100 image (32x32). 
To equalize the resolution of the images as model inputs, we stretched the images from (32x32) to (224x224). At the same time, we reduced the resolution of the other images to (224x224) due to their larger original resolution. We speculate that this process have affected the low accuracy of CIFAR-100.


In terms of the impact of post-processing, the results indicate that the post-processing led to an average improvement of 1.29\% in top-1 accuracy across 20 results, and up to 4.3\% increase with ResNet-152 under Oxford Pet. The findings demonstrate that the image post-processing of \sysname improves the model accuracy across different structures and datasets. However, our experiments did not provide clear guidelines for model- or dataset-specific post-processing techniques, and in four out of 20 results, the performance without post-processing was the best. 
Investigating the relationship between the performance and various post-processing techniques to find optimal post-processing for each model and dataset remains for our future work.


This experiment confirms the feasibility of \sysname, achieving classification accuracies up to 91.4\% (using ResNeXt101 on Caltech). Interestingly, we found that ResNeXt101 outperformed Swin-v2 by an average of 0.9\% across datasets, though the \sysname was initially designed with the Swin-v2. However, as the accuracy of the model depends on the hyperparameter setting, further research is necessary to explore the full potential of \sysname. Note that parameter optimization for each model was out of the scope.

\vspace{-0.8em}
\subsubsection{Diversity}
\vspace{-0.2em}
In this section, we examine the diversity between the real dataset and the \sysname dataset from two perspectives: 1) the impact of different prompt generators on reproducing single dataset (Oxford Pet), 2) the impact of \sysname's prompt generator on reproducing additional datasets.

Specifically, we verify whether the ChatGPT-based prompt generator of \sysname help generates a more diverse dataset in general cases. In the first perspective, we reproduced Oxford Pet into two different datasets - one by a prompt generator using fixed prompt (denoted as \sysname-Naive), and the other by a ChatGPT-based prompt generator (denoted as \sysname-ChatGPT). These datasets were then compared with the original dataset to measure the effects of the generators respectively. Furthermore, in the second perspective, we only use \sysname-ChatGPT to reproduce additional datasets, then compared them with the real ones to evaluate how generally diverse the \sysname datasets are. The average SSIM score is used as the measurement standard in every cases. SSIM~\cite{wang2004image} is a well-known method to assess the structural similarity between two images. We assumed that the lower average SSIM score over all pairs in a class indicates that the images in the class are composed of images with more different structures (i.e., diverse). 

\begin{figure}
\centering
\includegraphics[width=0.9\linewidth]{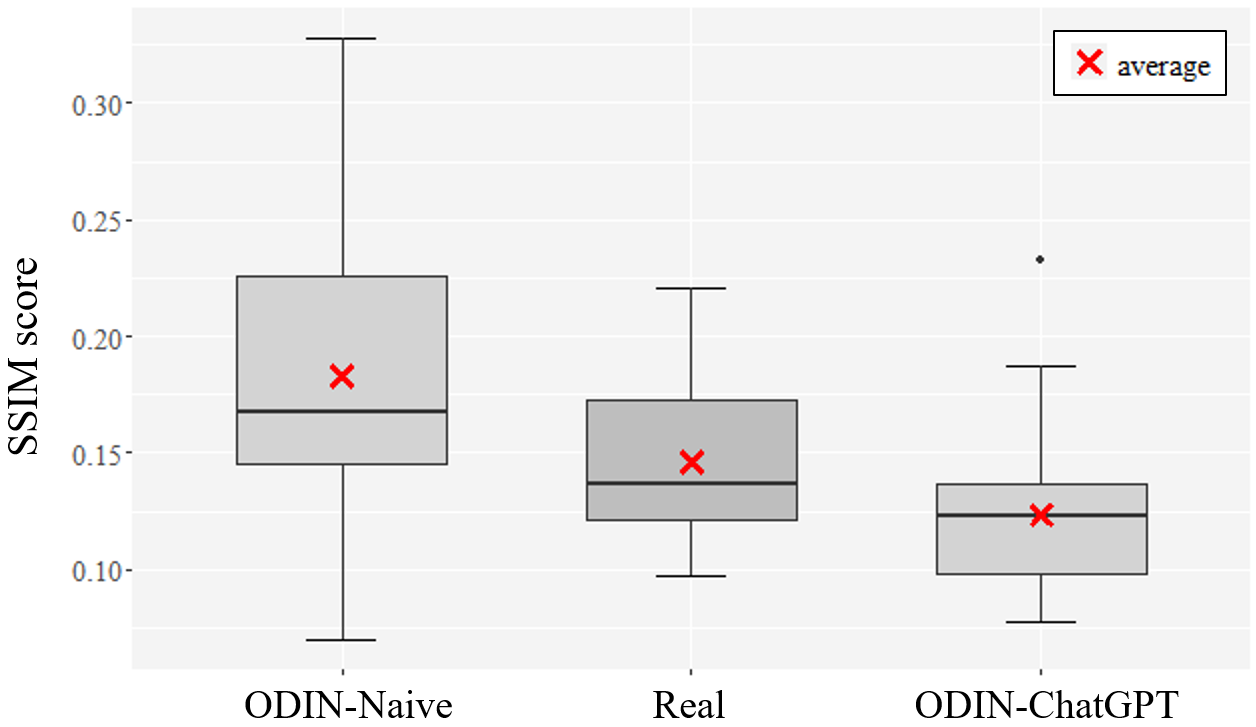}
\caption{Distribution of SSIM scores for Oxford Pet with its reproductions (\sysname-Naive, and \sysname-ChatGPT).}
\label{fig.oxford_ssim}
\end{figure}


\smallskip
\noindent \textbf{Different prompt generators on reproducing single dataset}:
Figure~\ref{fig.oxford_ssim} presents SSIM scores of images generated under classes of Oxford Pet, each based on the prompts of \sysname-Naive and \sysname-ChatGPT. The results show that the dataset with \sysname-ChatGPT looks more similar to the real ones than those with \sysname-Naive. 
The mean and variance of SSIM scores for each class are as follows: \sysname-Naive (M=0.18, SD=0.06), Real (M=0.15, SD=0.03), and \sysname-ChatGPT (M=0.12, SD=0.03). Dataset with \sysname-ChatGPT tends to have lower SSIM scores than the real dataset, probably because the increased prompt complexity generated more images drawn in unexpected compositions. 


\begin{table}
\centering
\small
\begin{tabular}{|c||c|c| }
\hline
 \multirow{2}{*}{\textbf{Datasets}} & \textbf{\sysname-ChatGPT} & \textbf{Real dataset}\\
 & SSIM (Mean/SD) & SSIM (Mean/SD) \\
\hline
 Oxford Pet & 0.12 / 0.03 & 0.15 / 0.06 \\
 Caltech  & 0.14 / 0.08 & 0.20 / 0.08 \\
 Indian Food & 0.08 / 0.02 & 0.11 / 0.03 \\
 CIFAR-100 & 0.10 / 0.05 &  0.40 / 0.06 \\      
 \hline
 \end{tabular}
 \caption{SSIM score with \sysname-ChatGPT and real dataset.}
 \vspace{-1em}
\label{section4-table-SSIM}
\end{table}

\smallskip
\noindent \textbf{\sysname-ChatGPT on reproducing additional datasets}:
We reproduced three additional datasets (Indian Food, CIFAR-100, and Caltech) with \sysname-ChatGPT, then compared SSIM scores with the real ones. Table~\ref{section4-table-SSIM} shows the results of all our runs. In the case of Indian Food, generated images with \sysname-ChatGPT obtained SSIM score of (M=0.08, SD=0.01), whereas the real dataset's SSIM score is in distribution of (M=0.11,SD=0.03).
A larger difference in SSIM scores between the real and the generated datasets was shown in the case of Caltech, with \sysname-ChatGPT's SSIM scores of (M=0.14, SD=0.08) and real dataset (M=0.20, SD=0.08).
This might be due to the presence of \textit{confusing} labels in Caltech that affects the prompt generator. For example, we asked ChatGPT to suggest a prompt with the word '\textit{airplanes}' (a class in Caltech), but it rather gave an answer including '\textit{helicopter}'. We assume that ChatGPT inadvertently judged helicopter as a kind of airplane.
Lastly, SSIM score of \sysname-ChatGPT in CIFAR-100 classes achieved (M=0.10, SD=0.05), while the real presenting (M=0.40, SD=0.06). 
Here, the resolution of the actual image is inherently low (32x32), which led to an underestimation in terms of structural similarity.


\vspace{-0.8em}
\subsubsection{Summary of findings} 
\vspace{-0.2em}
(1) A dataset generated by \sysname is useful to train various models on different datasets, however, the accuracy is slightly lower compared to the models trained on real images; (2) The diversity of images generated by \sysname is strongly correlated with the quality of the prompts; (3) While the model trained on \sysname dataset performs well for common resolution images, we found that it has limited performance in low resolution.

\subsection{Gap between the real and generated images}
We further investigate the reason for the differences in model accuracy between the real and the \sysname datasets. Our initial hypothesis was the presence of an invisible gap between the real and generated images. 
To verify this, we performed a binary classification to distinguish the real and the generated images using several model structures with similar parameter sizes (ResNeXt101: 83.5M, ViT-b-16: 86.6M, Swin-v2: 87.9M). We tested on different datasets used in \S4.1 without image post-processing.


\begin{table}
\centering
\small
    \begin{tabular}{|c||c|c|c|c|}
       \hline
         \multirow{2}{*}{\textbf{Models}}& \textbf{Oxford} & \multirow{2}{*}{\textbf{Caltech}} & \textbf{Indian} & \textbf{CIFAR}\\
          & \textbf{Pet} &  & \textbf{Food} &  \textbf{100}\\
        \hline
      ResNeXt101 & 94.8\% & \textbf{90.0\%} & 90.1\% & 99.0\%\\  
      ViT-b-16 & 91.9\% & 91.6\% & 91.9\% & 99.8\%\\  
      Swin-v2 & 95.7\% & \textbf{97.5\%} & 92.5\% & 98.5\% \\      
       \hline
    \end{tabular}
    \caption{Binary classification results (real vs. generated images).}
    \vspace{-2em}
    \label{tab:table_4.5}
\end{table}

Table~\ref{tab:table_4.5} shows the binary classification results for real and generated images across various models and datasets. 
We observed that the real and generated images were well distinguished regardless of the model structures, indicating the presence of invisible differences between them. 
ResNeXt101 and Swin-v2 in Caltech show the largest difference, implying that Swin-v2 performs better at identifying the generated and real images than ResNeXt101. 
To investigate the impact of such a difference on the model performance, we evaluated the performance of each model on both real and \sysname datasets under the same settings. 
ResNeXt101 achieved 89.6\% and Swin-v2 achieved 86.4\% of accuracy (see Table~\ref{section4-table:1}). In comparison, the accuracy for the real dataset was 96.5\% for ResNeXt101 and 94.8\% for Swin-v2. The performance difference was 6.9\% for ResNeXt101 and 8.4\% for Swin-v2. 
As Swin-v2 outperforms in distinguishing between real and generated images, we think this contributes to the larger performance difference between real and \sysname datasets.

This finding also explains why the post-processor was effective in improving the model performance, which reduces the gap between real and generated images.
Moreover, although the generated images exhibit clear differences from real images, the model performance on \sysname dataset still achieved reasonable accuracy. This is probably because the model was initially trained using a pre-trained ImageNet-1k checkpoint, which already has knowledge of real images, compensating for the difference between the real and generated images in the model performance.

\section{Discussion}
\subsection{Limitation}

\noindent \textbf{Locking in generalized pre-trained models}: \sysname addresses the dataset lock-in of traditional machine learning and mitigates the problem by utilizing generalized pre-trained models (i.e., Stable Diffusion and ChatGPT). However, it is still constrained by the knowledge boundaries of these models, which is unable to generate data beyond their knowledge. 
As such models continue to learn and expand their knowledge boundaries, \sysname can generate an unlimited amount of data within their knowledge range, providing the necessary data for personalized model training.

\smallskip
\noindent \textbf{Optimization for generated images}:
In \S2.1, we highlighted that previous research discussed an invisible gap between real and generated images~\cite{he2022synthetic, zhang2021datasetgan}. 
As we mentioned in \S4.2, we achieved 98.5\% accuracy in detecting real and generated images under CIFAR-100. It indicates that there are obvious differences between the generated and the real images.  
We conducted experiments in \S4.1 and found that the model trained on real images consistently outperformed those trained on generated images.
However, the hyperparameters may not be optimal for \sysname dataset. Given the difference between real and generated images, there is room for improvement in hyperparameter optimization for generated images, as well as in generating better prompts and image post-processing.

\subsection{Fueling AI adoption}
Data is often referred to as the \textit{fuel}~\cite{datafuel} that drives model training and boosts performance without directly impacting the model architecture. We discuss the potential of \sysname as the fuel for future adoption, highlighting its advantages in terms of high versatility and scalability.

\smallskip
\noindent \textbf{Within-domain versatility}:
Lack of data often discourages thorough experiments in certain domains. For example, in the case of human detection, deep learning model may detect humans accurately when they are facing forward or sideways, but perform poorly in angled photos taken from the ceiling~\cite{10.1145/3210240.3210348,10024513}. This is because there are simply insufficient data consisting of ceiling-perspective photos. \sysname addresses such issues straightforwardly without requiring additional manipulation or optimization of model structures, which ultimately facilitates practical utilization.


\smallskip
\noindent \textbf{Scalability to other domains}: 
\sysname also has an advantage in facilitating the adoption of AI in other domains where data collection is challenging.
Several domains in our lives suffer from data depletion due to the difficulties in collecting data, such as in a disaster or medical emergency. 
\sysname overcomes limited availability of data in such domains by generating synthetic data to train AI models for any situation in any domain. 
Imagine an earthquake where a building collapses under a huge fire breakout, and using traditional methods (e.g., infrared sensors) for survivor search impossible.
In this case, a camera with a built-in detection system pre-trained with the \sysname dataset under earthquake scenarios can be served as a viable substitution.
While taking a disaster as an initial example, we believe that the capability of \sysname (i.e., fueling data) opens up new opportunities for the adoption of AI in previously inaccessible areas.




\subsection{Intensifying user customization}

\sysname is designed to formulate on-demand dataset, in which the `customization' is the underlying pursuit. We discuss design flexibility and task extensibility, which have the potential to further enhance the level of customization.

\smallskip
\noindent \textbf{Design flexibility}:
We designed each component of \sysname to be modular and independent of the others so that they can be easily replaced by other models as needed. In particular, \sysname currently uses Stable Diffusion, which is one of the generalized pre-trained models. Although such models are referred to as ``generalized”, they are not always suitable for all use cases. However, there are opportunities to improve task performance by replacing them with more specialized models. For example in the classification of specific painting styles, replacing the Stable Diffusion with a model primarily trained for painting (e.g., Novel AI~\cite{novelai}) could potentially enhance the quality of the generated images. The flexibility in our design helps create a custom dataset that better meets user requirements and improves performance.

\begin{figure}
\centering
\includegraphics[width=0.94\linewidth]{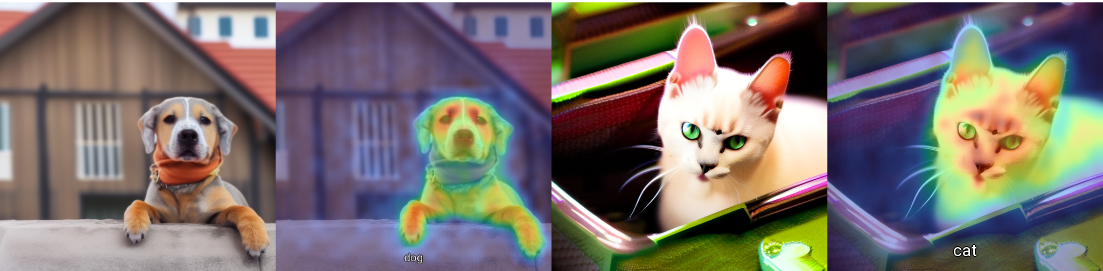}
\caption{Heatmaps presenting the parts of the generated images that correspond to the given label.}
\vspace{-1em}
\label{fig.pixel_coordinate}
\end{figure}

\smallskip
\noindent \textbf{Task extensibility}:
\sysname can expand the scope of the task by adding a few more components to its design beyond replacing them. Figure~\ref{fig.pixel_coordinate} shows heatmaps of the parts of the generated images that correspond to given labels. In this manner, it is possible to create a dataset for image recognition or image segmentation tasks by providing the bounding box of a given label or its pixel coordinates along with the entire image. This functionality allows users to create a custom dataset tailored to specific tasks and requirements.

\section{Conclusion}
We address the problem of dataset lock-in by dynamically formulating on-demand dataset. We propose \sysname, a system that consists of 3 main modules: a prompt generator, a txt2img generator, and an image post-processor. We evaluate \sysname in terms of accuracy and diversity. Our results demonstrate the potential of \sysname to enable models to learn new knowledge beyond the training dataset. We believe \sysname represents a step forward from the zero-shot learning and has the potential to change the way we approach deep learning problems.

{\small
\bibliographystyle{ieee_fullname}
\bibliography{egbib}
}



\section*{Supplementary material}

# ***Supplementary Material***
# ODIN: <u>O</u>n-demand <u>D</u>ata Formulation to M<u>I</u>tigate Dataset Lock-i<u>N</u>

## 1. Implementation of ODIN

Our current prototype implementation utilizes ChatGPT API from OpenAI to receive prompt responses. While our design is capable of full automation, we still need to implement further functionality to handle unexpected responses from ChatGPT. To demonstrate that ODIN works in practice, we share our Colab code that includes an example implementation in ODIN-Colab[1].

The final implementation of ODIN utilizes ChatGPT for the prompt generator. In §3.2 of the paper, we also discussed an alternative approach where the original image is captioned and used as the prompt. Figure 1 provides more details, including original images, blip-v2-based image captioning, replacing coarse subject to the label as the prompt, and generated images.

Lastly, we applied various noise techniques in our image post-processing to reduce the feature distance between the real and the generated images. Figure 2 shows the noise techniques we used for the image post-processor.

## 2. Advanced approach: Prompt diversification

In §3 of the paper, we highlighted the importance of diverse prompts. Here, we suggest a novel approach to strengthen the diversity of the prompts by leveraging the pre-trained word2vec [2] model on top of the existing prompt generator. Our key idea is to diversify the properties of the target label in the prompt. We adopted the presence of life as a baseline property, and added modifiers accordingly. Algorithm 1 shows the process in step-by-step pseudo-code.

Firstly, we created lists $P$ and $N$, consisting of words that describe living and non-living things, respectively. Next, we measured the cosine distance between each list and the label to determine which group the label is closest to. We then added modifiers along with the label to diversify its properties. For instance, taking the word 'lion' as an input, the overall prediction would show $avg(V_p)$ greater than $avg(V_n)$, thus we create phrases such as 'female lion', 'young lion', and 'sick lion'. This approach is layered on top of ODIN's prompt generator, and the labels of the generated prompts are replaced accordingly. We expect improve-

---

**Algorithm 1** Word2vec-based prompt diversification

    **Input**: $X$                                      ▷ X: Label
    **Output**: $\mathcal{X}_{phrase}$       ▷ $\mathcal{X}_{phrase}$: Phrase containing X
1: **function** GETVECTOR($word$)
2:     **return** word vector on pre-trained Word2Vec model
3: **end function**
4: $P \leftarrow$ a list of words that describe living things
       (e.g., animate, animal, plant)
5: $N \leftarrow$ a list of words that describe non-living things
       (e.g, inanimate, object, man-made)
6: **for** item $i$ in P **do**
7:     **for** item $j$ in N **do**
8:         $V_p \leftarrow$ Sum of cosine similarities between
            GetVector($P_i$) and GetVector($X$)
9:         $V_n \leftarrow$ Sum of cosine similarities between
            GetVector($N_j$) and GetVector($X$)
10:     **end for**
11: **end for**
12: **if** $avg(V_p) > avg(V_n)$ **then**
13:     $\mathcal{X}_{phrase} \leftarrow$ Phrase containing words that modify
           living things (e.g., female, young, sick)
14: **else**
15:     $\mathcal{X}_{phrase} \leftarrow$ Phrase containing words that modify
            non-living things (e.g., broken, red(color))
16: **end if**
17: **return** $\mathcal{X}_{phrase}$

---

ments in the diversity of generated images since this approach adds various extra information to the label.

## 3. Extended experiment: Diversity

In §4 of the paper, we evaluated the dataset based on the structural similarity of images (SSIM). In this supplementary material, we included the graphical results for the color of images, which provide additional insights. We used four datasets (including Oxford Pet, Indian Food, Caltech, and CIFAR-100) that were reproduced by ODIN and their original images. As for the color aspect, we measure the '*colorfulness*' of an individual image using the proposed measurement method in [1], and present the distribution of '*col-*

---
[1] https://tinyurl.com/ODIN-Colab

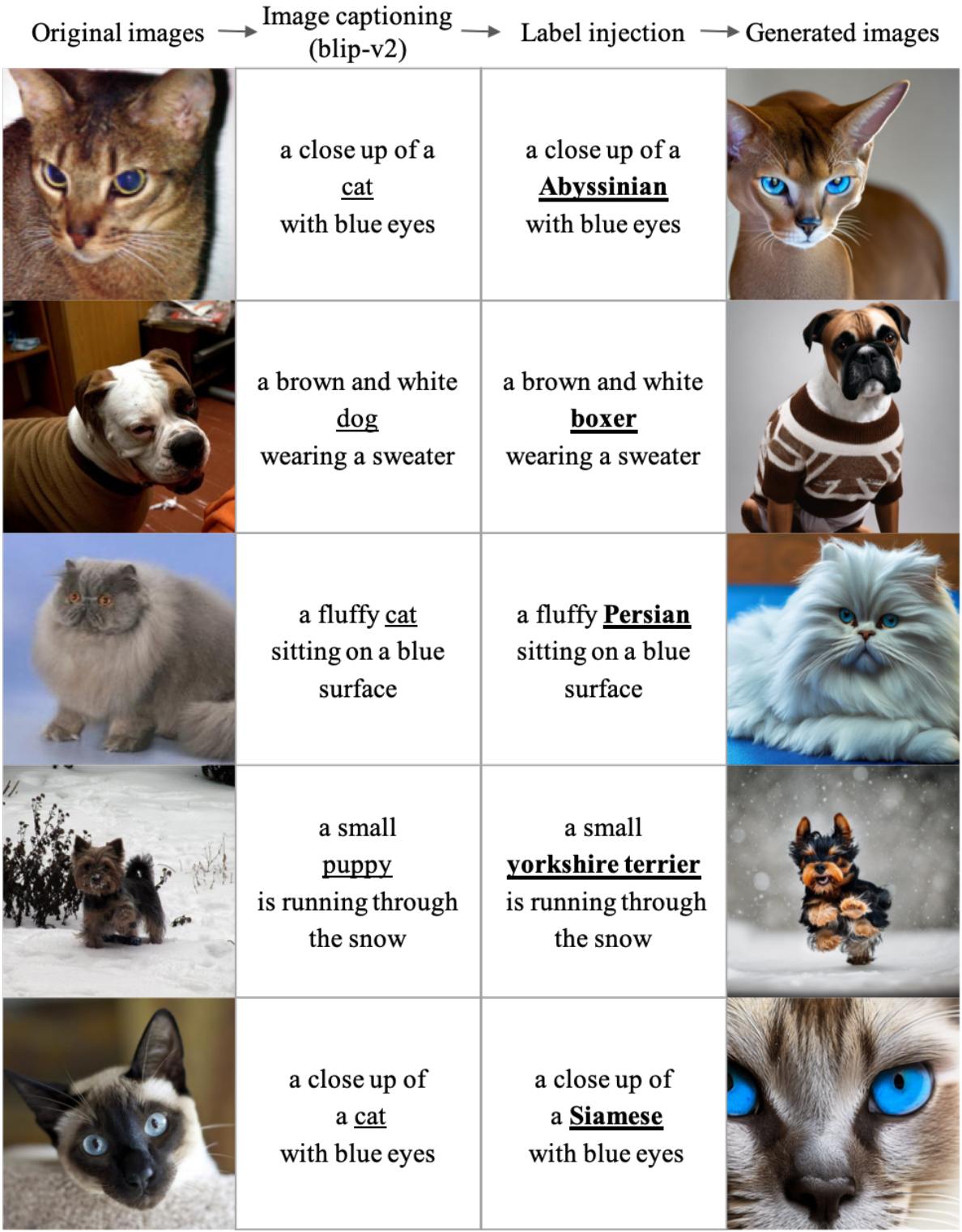

Figure 1. Prompt and image generation based on the image captioning (blip-v2); (1) original images are captioned, (2) the coarse subject of the caption is replaced with the label as the prompt, and (3) the image data is generated from the prompt.

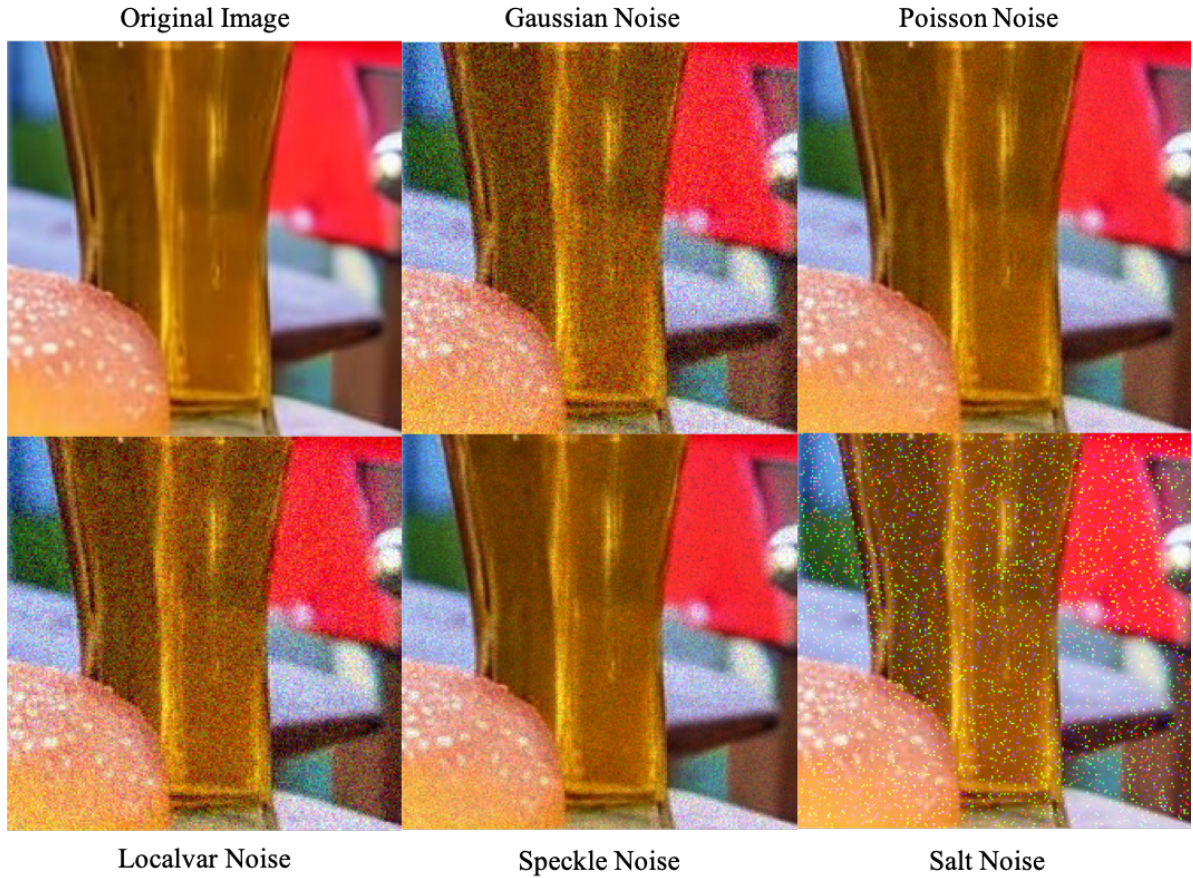

Figure 2. A original image (upper left) and five noised images with different noise techniques (center cropped, 768 to 224).

*orfulness*' in both the ODIN dataset and the real dataset. All images were resized to 224x224 for the measurement. The results are shown in Figure 3. We could not observe significant differences between the two datasets, yet ODIN dataset has slightly more diverse '*colorfulness*'.

## 4. Future work: Handling unexpected prompt

As shown in Figures 4 and 5, unexpected prompts (discussed in §4.1.2 of the paper) may degrade the performance of the model trained by ODIN dataset as they lead to the generation of incorrect image data. It is possible to filter out incorrect images after complete generation, however, this approach is computationally inefficient. At this point, complementary research regarding small-scale language models that can determine the complexity or appropriateness of the prompts are needed for future work.

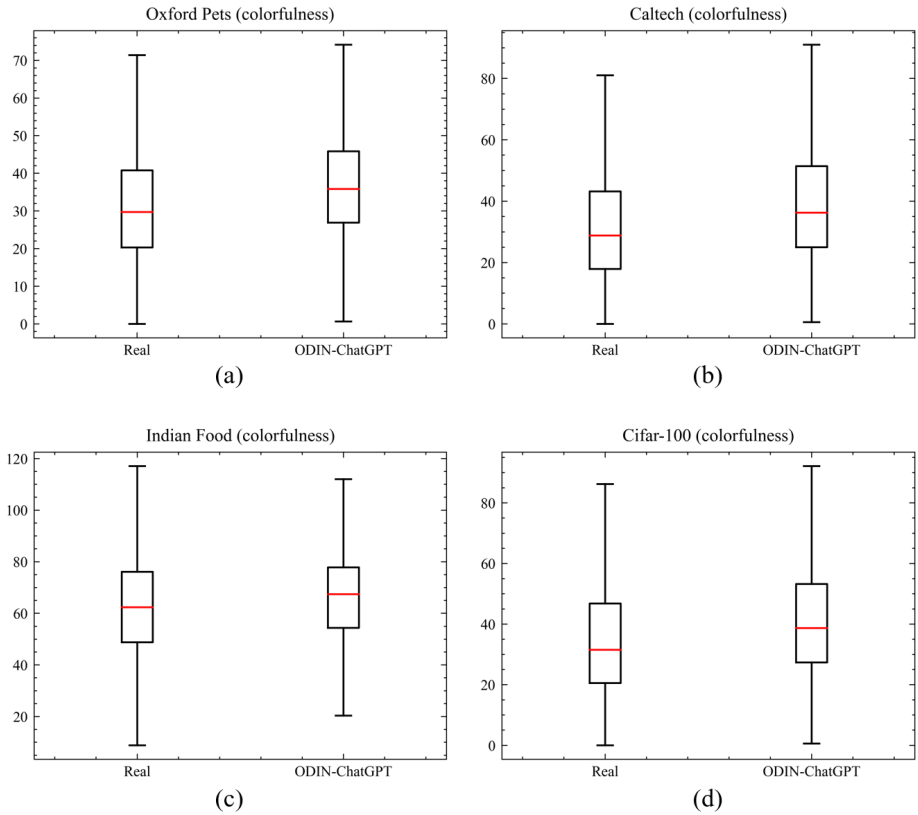

Figure 3. The results of '*colorfulness*' on real dataset and ODIN dataset; (a) Oxford Pets, (b) Caltech, (c) Indian Food, and (d) Cifar-100.

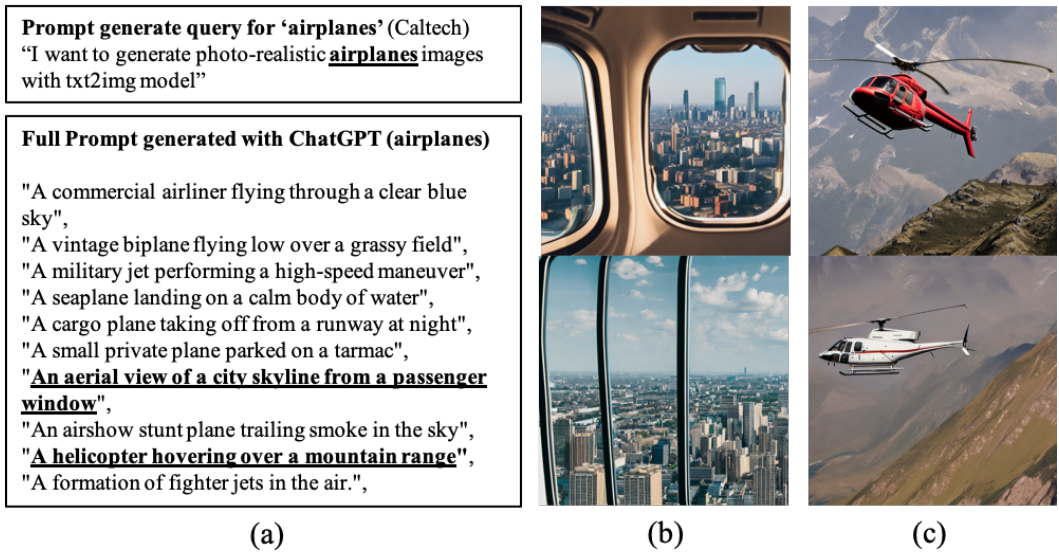

Figure 4. (a) The query ('airplanes' class) and the responses with ChatGPT API. (b and c) Sample generated images from two unexpected prompts, 'An aerial view of a city skyline from a passenger window' and 'A helicopter hovering over a mountain rage', respectively.

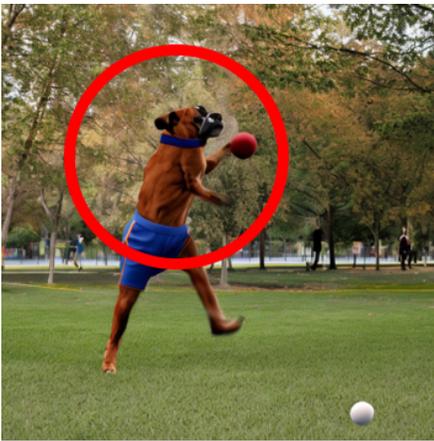 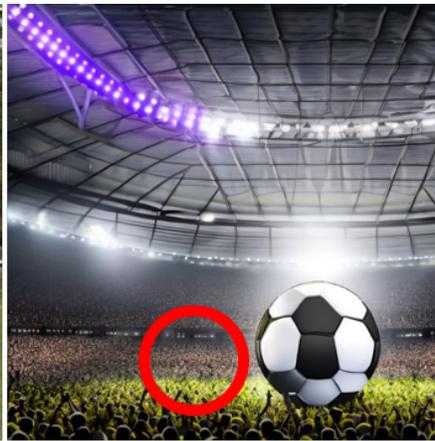 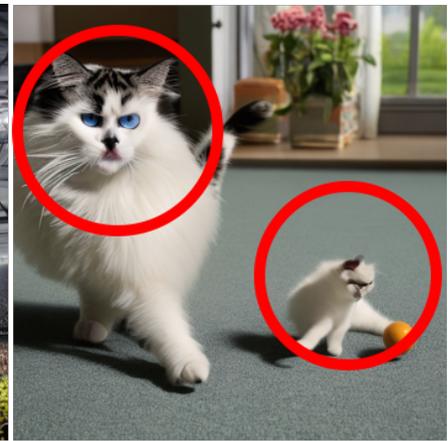

Figure 5. Example images which have unexpected composition, derived by complex prompts from ChatGPT



\end{document}